\documentclass[pmlr]{jmlr}

\RequirePackage{graphicx}
\usepackage{multirow} 
\usepackage{subcaption}
\usepackage[skip=6pt]{caption}
 \usepackage{booktabs}
\usepackage{makecell} 
\usepackage{longtable}
\usepackage[utf8]{inputenc}
\usepackage{pifont}
\usepackage{newunicodechar}
\newunicodechar{✓}{\ding{51}}
\newunicodechar{✗}{\ding{55}}

\makeatletter
\def\set@curr@file#1{\def\@curr@file{#1}} 
\makeatother
\usepackage[load-configurations=version-1]{siunitx} 

\theorembodyfont{\upshape}
\theoremheaderfont{\scshape}
\theorempostheader{:}
\theoremsep{\newline}

\jmlrvolume{252}
\jmlryear{2024}
\jmlrworkshop{Machine Learning for Healthcare}

\title[Selective Fine-tuning with LLM-labeled Data]{Selective Fine-tuning on LLM-labeled Data May Reduce Reliance on Human Annotation: A Case Study Using Schedule-of-Event Table Detection}

\author{\Name{Bhawesh Kumar}
       \Email{bhaweshk@verily.com}\\ 
       \addr Verily Life Sciences\\
       269 East Grand Avenue, South San Francisco\\
       \AND
       \Name{Jonathan Amar}
       \Email{jonathanamar@verily.com}\\ 
       \addr Verily Life Sciences\\
       269 East Grand Avenue, South San Francisco\\
       \AND
       \Name{Eric Yang}
       \Email{eryang@verily.com}\\ 
       \addr Verily Life Sciences\\
       269 East Grand Avenue, South San Francisco\\
       \AND
       \Name{Nan Li}
       \Email{notanumber@verily.com}\\ 
       \addr Verily Life Sciences\\
       269 East Grand Avenue, South San Francisco\\
       \AND
       \Name{Yugang Jia}
       \Email{yugang@verily.com}\\ 
       \addr Verily Life Sciences\\
       269 East Grand Avenue, South San Francisco\\
       }
\begin{document}

\maketitle

\begin{abstract}
  Large Language Models (LLMs) have demonstrated their efficacy across a broad spectrum of tasks in healthcare applications. However, often LLMs need to be fine-tuned on task-specific expert-annotated data to achieve optimal performance, which can be expensive and time consuming. In this study, we fine-tune PaLM-2 (\cite{anil2023palm}) with parameter efficient fine-tuning (PEFT) using noisy labels obtained from gemini-pro 1.0 (\cite{geminiteam2024gemini}) for the detection of Schedule-of-Event (SoE) tables, which specify care plan in clinical trial protocols. We introduce a filtering mechanism to select high-confidence labels for this table classification task, thereby reducing the noise in the auto-generated labels. We find that the fine-tuned PaLM-2 with filtered labels outperforms Gemini Pro 1.0 and other LLMs on this task and achieves performance close to PaLM-2 fine-tuned on non-expert human annotations. We also explore supplementing LLM-generated labels with a limited number of human annotations for challenging cases. Our results show that leveraging LLM-generated labels, coupled with strategic filtering and selective human annotation, can be a viable and cost-effective strategy for improving LLM performance on specialized tasks, especially in domains where expert annotations are scarce, expensive, or time-consuming to obtain.

\end{abstract}

\section{Introduction}

Large Language Models (LLMs) have been found to be useful across diverse tasks like natural language understanding and generation, question-answering, summarization, programming, and creative arts (\cite{chen2021evaluating, radford2018improving, radford2019language, ramesh2021zeroshot}).  LLMs are particularly promising in specialized fields such as healthcare, where they can significantly enhance clinical decision-making, patient care, drug discovery, and the management and utilization of medical data (\cite{singhal2023large, ingraham2023illuminating, doi:10.1056/AIoa2300138, tu2024towards, sharma2024assessment}). However, the successful application of LLMs in specialized domains frequently depends on their ability to process and understand complex, domain-specific structured and unstructured content, which often requires fine-tuning the models with data annotated by experts (\cite{van2023exploration}).  This necessity presents considerable challenges, primarily due to the scarcity, high cost, and substantial time required to acquire expert annotations in fields like healthcare. In response to these challenges, our work investigates the potential of LLM-generated labels for fine-tuning purposes, with a specific case-study on identifying Schedule-of-Event (SoE) tables in clinical trial protocols. The accurate identification of SoE tables, which outlines plan-of-care in clinical trials (also see appendix \ref{soe_table} for more details on SoE tables), plays a pivotal role in the digitization of clinical trial protocols which we briefly describe below.

\subsection{Brief Introduction to Clinical Trial Protocols and Digitization}

Clinical trials are the backbone of medical research. However, the traditional conduct of clinical trials is fraught with inefficiencies at various stages including patient recruitment, follow-ups, data acquisition and handling (\cite{inan2020digitizing, marquis2019technology}). Clinical trials rely heavily on manual processes, leading to time-consuming, expensive, and error-prone workflows. The inefficiencies pose challenges to all stakeholders involved in the trial and also slow down the pace of medical research (\cite{getz2017trial, jones2016changing}). 

Clinical trial protocols are foundational documents in the trials, outlining the detailed methodologies, objectives, and care plans that guide the conduct of studies in accordance with regulatory, ethical, and scientific standards. These protocols include critical components such as the Schedule of Events (SoE) table, which details the plan of care for participants, including visits for screening, treatment, and follow-up phases, along with the assessments, treatments, and data collection scheduled for these visits. The digitization of clinical trial protocols refers to the process of converting these detailed and often voluminous paper-based documents into accurate digital workflows (\cite{verily_viewpoint_2023, rosa2021using, inan2020digitizing}). This transformation is not just a matter of changing the medium but involves the systematic identification, classification and ultimately extraction of key elements within the protocols, such as SoE tables, to ensure they are accurately captured and can be effectively managed and analyzed in a digital system (\cite{inan2020digitizing}). Correctly identifying these tables, which can vary significantly in formatting, terminology, and layout across different protocols, poses a significant challenge (refer to appendix \ref{soe_table} for more details on SoE tables as well as examples.) However, the accurate classification of such tables are crucial for any automated protocol digitization workflow; an undetected SoE table can lead to an incomplete care plan, while a misclassified table introduces erroneous information into the system, underscoring the paramount importance of reliability in this process.

\subsection{Improving Domain-Specific LLM Performance with Synthetic Labels}

\subsubsection{Modeling SoE Detection with LLMs}
We model the problem of SoE table classification as a binary classification problem and use an LLM (PaLM-2) for accurate classification of SoE tables. To improve PaLM-2's ability to accurately classify SoE tables with fine-tuning, we use gemini-pro 1.0 to auto-generate training labels for fine-tuning task. This strategy aims to address the challenges of acquiring expert annotations by leveraging the capabilities of LLMs to produce high-quality, task-specific data.

\subsubsection{Fine-Tuning LLM with LLM-Generated Labels}\label{ft_llm_human}
The labels obtained from LLMs for specialized tasks like SoE table classification can be quite noisy. Thus, for a fine-tuning task to succeed, we need to remove potentially incorrect labels from auto-generated labels. For our specific task of SoE table classification, we use the consensus in gemini-pro 1.0 model inference across dual data representations of tables – JSON and text representations – to reduce noise in the training dataset for PaLM-2. Specifically, we fine-tune PaLM-2 models on only those LLM labels, where the JSON and text based inferences of the tables are identical for the label generating LLM (gemini-pro 1.0 in this case.)

The JSON representation of the table, which represents each of the table columns as a dictionary with the key being the row number and the value being the cell value for that column, preserves the structural details of the table. In contrast, the text representation encompasses not only the contents within the table but also all surrounding text on the page, including footnotes, titles, and any other textual content. This comprehensive capture of page content provides a fuller context and valuable redundancy for our inference process, improving the model's ability to accurately interpret and classify the tables.

We find that enhancing the quality of the auto-generated dataset for fine-tuning PaLM-2 leads to substantial improvement in fine-tuned model performance. The PaLM-2 model trained with these subset of LLM generated labels outperforms both the baseline PaLM-2 and gemini-pro 1.0 on SoE detection task (see table \ref{tab:test_set_results}). Remarkably, the fine-tuned PaLM-2 model achieves performance levels close to those obtained with human-annotated labels, showcasing the effectiveness of using LLM-generated labels for domain-specific tasks, particularly in settings where expert annotations are sparse. Finally, we explore a hybrid approach where a limited number of low-confidence LLM-generated labels, as identified by our filtering mechanism (i.e., instances where JSON and text-based inferences disagree), are selectively replaced with human annotations. This strategy aims to leverage the efficiency of LLM-generated labels while mitigating potential concerns regarding biases that might arise from discarding labels. Although the performance of the models fine-tuned with human and gemini annotations with the filtering mechanism is already comparable, this hybrid approach utilizing only 10\% of human labeling further narrows the gap with the PaLM-2 model trained with human labels as seen in table \ref{tab:test_set_results}.

\subsection{Generalizable Insights about Machine Learning in the Context of Healthcare}

Our work on SoE table classification, a highly specialized healthcare task, provides valuable insights into the potential of Large Language Models (LLMs) for advancing machine learning applications in healthcare. Our work demonstrates that LLM-generated labels, when combined with filtering and selective human validation, may offer a scalable and cost-effective alternative to traditional expert annotation for fine-tuning domain-specific models. This has significant implications for healthcare, where access to expert annotations is often limited due to cost, time constraints, and the limited availability of experts.

We summarize the key insights below:

\begin{itemize}
    \item \textbf{LLMs may be effectively applied to highly specialized healthcare tasks and have the potential to transform manual workflows:}  Our results show that even for tasks that require a deep understanding of domain-specific terminology and concepts, like SoE table classification, LLMs can potentially achieve strong performance, especially after fine-tuning. This adaptability of LLMs highlights their potential to automate and streamline traditionally manual and error-prone processes in healthcare. 
    \item \textbf{LLM-generated labels offer a promising solution for addressing annotation challenges:} Leveraging LLMs to generate labels presents a scalable and cost-effective alternative to traditional manual and expert-based annotation, potentially accelerating the development of ML models in healthcare.
    \item \textbf{Strategic filtering and human validation of LLM generated labels are crucial for optimizing performance:} Our filtering mechanism, based on dual data representations, significantly improves the quality of LLM-generated labels and subsequent fine-tuned model performance. Our hybrid labeling strategy shows that selective human annotation on difficult examples may further improve performance and bridge the gap with models fine-tuned entirely on human labels.
\end{itemize}

\section{Related Work}


\subsection{Large Language Models in Healthcare}
Recent advances in natural language processing (NLP) and machine learning have significantly enhanced the potential for integrating these technologies into various aspects of healthcare, including clinical decision-making, patient care, drug discovery, and medical information management. A wealth of studies have underscored the capabilities of Large Language Models (LLMs) in performing crucial NLP tasks in healthcare and medicine, such as extracting medical information, summarizing patient information, facilitating automated diagnosis, and even passing board certification exams in specialty medicines (\cite{liu2021ai, shay2024could, van2023clinical, ingraham2023illuminating, tu2024towards, doi:10.1056/AIoa2300138}). These applications highlight the potentially transformative impact LLMs could have on healthcare.

In the context of clinical trials, LLMs have been utilized to parse and understand interventions and findings from randomized control trials (\cite{wadhwa2023jointly}), and to assist in patient-matching for clinical trials by analyzing electronic health records (EHRs) alongside clinical trial documentation (\cite{yuan2023llm}). Previous research has also studied the problem of automated identification of specific elements from Schedule-of-Event (SoE) tables, such as detailed activity information, employing a human-in-the-loop approach to ensure accuracy and relevance (\cite{dhuliawala2018happens}). 

These emerging applications not only underscore the versatility of LLMs in managing diverse and complex healthcare datasets but also illustrate a pivotal challenge: the dependency on extensive, expert-annotated datasets for fine-tuning and evaluating LLMs in specialized tasks. This has led to a growing interest in automated label generation techniques and the exploration of fine-tuning and testing of LLMs with these synthetic labels.

\subsection{Model Training on Synthetic Dataset}
Successes of generative models in various tasks have spurred research into leveraging these models to augment real data for model fine-tuning and validation. \cite{DBLP:journals/corr/abs-1911-02888}, for example, use class-conditional GAN generated image for training model for image classification tasks. \cite{he2023synthetic} study the potential of synthetic data in zero-shot and few-shot classification using CLIP model (\cite{radford2021learning}. Recently, researchers have also used LLMs to augment data for various classification tasks. \cite{meng2022generating} use a pre-trained language model to generate samples by prompting it with real data and using the generated data for fine-tuning a BERT model. To control the quality of samples, they use log-probability of generated samples for filtering poor quality auto-generated samples. \cite{yoo-etal-2021-gpt3mix-leveraging} use randomly sampled existing data samples to condition models to generate new samples, while using token probability corresponding to the label-classes to obtain soft probability for these generated samples. These soft probabilities for synthetic samples are used to train BERT-style models for classification. One of the recent studies by \cite{li2023synthetic} has tried to understand when synthetic data can be helpful in successful model training. They find that synthetic data is less effective when a classification task is subjective or when a specific instance of data to be classified is subjective as measured by agreement amongst annotators. In the healthcare domain, \cite{feder2024causal} have explored counterfactual data augmentation for improved LLM generalization showing promising result on text classfication tasks. 

Building on these previous research, our study employs LLM generated labels for fine-tuning another LLM for table classification task in a highly specialized context, specifically, Schedule-of-Event table classification in clinical trial protocols. Distinct from previous research, which often relies on standard benchmarks or datasets for generating synthetic data, our work showcases a novel application of synthetic labels for fine-tuning in domains where expert annotation is expensive and challenging to obtain. Furthermore, we offer a detailed comparison between models fine-tuned on LLM-generated labels versus those fine-tuned on labels annotated by human experts. Our approach also introduces an innovative label filtering mechanism that utilizes dual data representations of tables for removing potentially noisy synthetic labels and allowing for selective annotation by human annotators. Finally, our approach doesn’t require access to logits for tokens and can be applied even when working with LLMs through black-box API access.

\section{Methods}

\subsection{Problem Set-up}

\begin{figure}[htb]
  \centering
  \includegraphics[width=\linewidth]{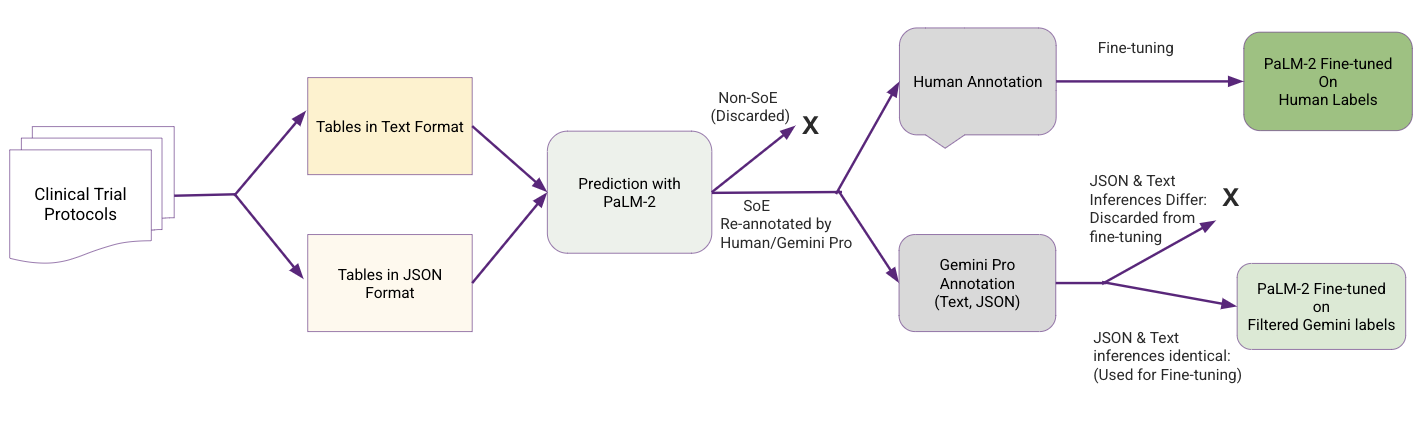} 
  \caption{\textbf{Schematic of the fine-tuning process.} This figure illustrates our approach to fine-tuning PaLM-2 for SoE table classification using LLM-generated labels. A base PaLM-2 model is first applied to identify potential SoE tables in 408 unlabelled protocols. Only predicted SoE tables are then annotated by both Gemini Pro (using JSON and text representations) and human annotators. Our results (table \ref{tab:test_set_results}) show that our filtering approach (where we discard Gemini-pro labels with different inferences for JSON and text view before fine-tuning) achieves performance close to that of fine-tuning on human labels, highlighting the potential of LLM-generated labels for specialized healthcare tasks.}
  \label{fig:flow}
\end{figure}
We frame the Schedule-of-Event table detection in a clinical protocol as a binary classification task of correctly classifying a table as a SoE table or a non-SoE table. Specifically, for each clinical trial protocol, the goal is to classify all the tables present inside that protocol as SoE or non-SoE table. We define a table as a SoE table when our in-house protocol digitization specialists label it as a SoE table. The goal is to achieve a very high level of precision and recall on table classification task. Since we don’t expect the model to be perfect in classification of the table, the classification algorithm is supposed to be used for reducing the annotator's burden of going through every page in a long protocol. All protocols digitized through this semi-automated approach with human-in-the-loop goes through stringent review and quality checks to ensure accuracy.

\subsection{Dataset \& Models}
\subsubsection{Training and Test Set}\label{train_test_set}
Our data set consists of a total of 499 clinical trial protocols of which 91 are expert-labeled by a team of five protocol digitization specialists and are used as the test set in our experiments. These 91 test protocols have a total of 3019 tables with 411 SoE tables (13.6\%) and 2608 non-SoE tables (86.4\%.) These expert-digitizers are specifically trained to manually label and digitize the clinical protocol for our in-house clinical trial management system (CTMS) software and the labeling process and digitization requires significant domain knowledge, time and effort.  We take the expert annotations as ground truth for all experiments. The subset of 408 protocols that don’t have any expert-labels are used for the fine-tuning tasks (see figure \ref{fig:flow}). Of the 408 protocols, we randomly select 300 as training set, 18 as validation set and 90 as test set for model fine-tuning task. Note that we use a separate expert-labeled test data set of 91 protocols for all model evaluations distinct from the one used during fine-tuning. These 499 protocols in our experiments span a diverse set of clinical trials across pharmaceutical companies, academic organizations, hospitals, and government organizations. 

\subsubsection{Models}
We use GPT-4 API (gpt-4-0613) (\cite{openai2024gpt4}), PaLM-2 (text-bison@001 on GCP) (\cite{anil2023palm}), and gemini-pro 1.0 (\cite{geminiteam2024gemini}) for our inference tasks. The base models (without any fine-tuning) serve as the baselines. We use the PaLM-2 model for all the fine-tuning experiments. We note that gemini-pro 1.0 and GPT-4 models have been reported as having substantially better performance on LLM benchmarks than PaLM-2 (\cite{geminiteam2024gemini}). The gemini-pro 1.0 model is not available for fine-tuning as of this writing.

\subsection{Selective Human and LLM Annotation} \label{selective}
For our fine-tuning task, we selectively collect human and gemini-pro 1.0 annotations on previously mentioned 408 protocols. The annotation is done by a team of six non-experts annotators and they can mark complex cases for review by the expert annotators as well as directly ask about any specific annotation from an expert. 

We show the fine-tuning process workflow in figure \ref{fig:flow}. We first do inference of all 408 unlabelled protocols with the PaLM-2 model. On a subset of 60 protocols, we manually go through PaLM-2 model prediction with the help of experts to find specific patterns in incorrect model prediction. We find that the base PaLM-2 has a very high recall but also a very high false positive rate and often predicts trivial cases of non-SoE tables as SoE. Thus, we only obtain human and gemini-pro 1.0 annotations on tables identified as SoE by the base PaLM-2 model (around 25\% of all tables.) This selective annotation allows us to keep the size of annotation tasks manageable (by reducing the task to one-fourth), while also helps us over-sample the SoE table examples for fine-tuning. Additionally, this approach allows annotators to concentrate their efforts on more ambiguous cases potentially leading to higher quality annotations since annotators can spend more time on each item. 

We summarize the results of non-expert and gemini-pro 1.0 based annotations in table \ref{tab:annotation_summary}. The train, validation, and test set for fine-tuning consist of 300, 18, and 90 protocols respectively. The number of SoE and non-SoE table annotation counts for non-expert and gemini-pro 1.0 annotations differ (since neither the non-expert human annotators nor gemini-pro 1.0 are perfect at identifying SoE tables and they may annotate a specific table differently), but total table counts are the same across various data splits.

\begin{table}[h]
\centering
\caption{Summary of Annotations}
\begin{tabular}{lccc}
\toprule
\textbf{Annotation Type} & \textbf{Train Set} & \textbf{Validation Set} & \textbf{Test Set} \\
\midrule
Non-Expert & 1536 SoE, & 53 SoE, & 383 SoE, \\
           & 1264 Non-SoE  & 74 Non-SoE & 413 Non-SoE \\
\midrule
Gemini-pro & 1748 SoE, & 64 SoE, & 490 SoE, \\
           & 1052 Non-SoE  & 63 Non-SoE & 306 Non-SoE \\
\bottomrule
\end{tabular}
\label{tab:annotation_summary}
\end{table}

We emphasize that we do not use expert annotators directly for labeling tasks. However, the annotators do have some previous experience with annotation for SoE tables and they also have access to expert annotators for any annotation they need help with. Additionally, they can choose to not annotate a table and leave it as “Do not know”. These are later annotated by an expert. Despite access to experts, non-expert annotations can be noisy due to variation in skills among the non-expert annotators. On random overlapping sets of 50 annotation, the average inter-rater agreement among non-expert annotators is 81.2\%. We note that all annotations are collected only on tables predicted as SoE by base PaLM-2 models as described previously. Thus, the inter-rate agreement is only on a portion of all tables present in the protocols that are annotated by human experts. For tables with multiple annotations, we choose the first annotation collected for that table.

Finally, to get a sense of the alignment between between Gemini-pro and the non-expert labels, we looked at the overlap between these two sets of annotations. We found that 89.6\% of the annotations between human labelers and gemini-pro are identical. We show the agreement matrix for human and gemini-pro labels in table \ref{tab:gemini_overlap}. We can see that majority of the disagreement are cases where gemini-pro classifies a table as SoE, but human raters don't. This is in-line with a high recall and low precision observed for gemini-pro model as seen in table \ref{tab:test_set_results} on the test set.

\begin{table}[h]
\centering
\caption{Agreement between Gemini Pro and Non-Expert Labelers}
\begin{tabular}{lccc}
\toprule
 &  & \multicolumn{2}{c}{\textbf{Non-Expert Labelers}} \\ 
\cmidrule(lr){3-4}
 &  & \textbf{SoE} & \textbf{Non-SoE} \\
\midrule
\multirow{2}{*}{\textbf{Gemini Pro}} & \textbf{SoE} & 1497  & 251  \\
 & \textbf{Non-SoE} & 39   & 1013  \\ 
\midrule
\textbf{Overall Agreement} & \textbf{89.6\%} \\ 
\bottomrule
\end{tabular}
\label{tab:gemini_overlap}
\end{table}

\subsection{Experiments}
We use PaLM-2, gemini-pro 1.0 and GPT-4 (gpt-4-0613) in our experiments. As described in section \ref{ft_llm_human}, we use JSON and text representations of a table for inference. We use camelot (0.11.0) (\cite{camelot}) for extracting JSON representations and pdfminer.six (20231228) (\cite{pdfminersix}) for text extraction. The model is asked to respond with “YES” or “NO” corresponding to the prompts (see Appendix \ref{prompts} for the prompts) for these JSON and text representations of the table. If either of the JSON or text inference output corresponding to a table is “YES”, we classify the table as SoE. This conservative approach for classification of SoE leads to higher false positives, but those are more easier to rectify in our digitization workflow than missed SoE tables, which can lead to missed plan-of-care and cause expensive manual corrections at later steps in the protocol digitization process. 

For fine-tuning the PaLM-2 model, we use the 408 protocols as previously described in section \ref{train_test_set}. We fine-tune all PaLM-2 models on table annotations obtained from 300 protocols and use 18 protocols for validation and the rest 90 protocols as test set for our fine-tuning process. We fine-tune three sets of models with gemini-pro 1.0 generated annotations and one model with human annotations. The first gemini-pro based fine-tuned model uses all 2800 table annotations from 300 protocols, while the second removes all “noisy labels” from fine-tuning. Specifically, for the second fine-tuning experiment with gemini-pro annotations, we remove all samples from training where the JSON and text annotations of gemini-pro are not identical. This reduces the set of table annotations to 2518 tables in the training set. Finally, for the third fine-tuned model, instead of discarding the 282 samples (having different text and JSON inferences), we replace them with human annotations. All models are fine-tuned for 300 epochs with learning rate multiplier of 1, early stopping set to True, and an evaluation interval of 10 epochs with Google Cloud Vertex AI fine-tuning pipeline which uses Parameter Efficient Fine-tuning (PEFT). We track the model training through a tensorboard instance.



\section{Results}
We evaluate models on a comprehensive set of metrics to assess the effectiveness of each model in the context of clinical trial protocol digitization. The models are benchmarked based on recall, precision, F-1 score, and accuracy. We additionally measure model performance at various precision threshold as well as on the percentage of protocols achieving 100\% recall and precision (refer appendix \ref{additional_metrics}), which are critical for the practical deployment of the automated digitization pipeline. To estimate the uncertainty in our performance metrics, we use bootstrapping with 10,000 replications. This involves repeatedly resampling the test dataset with replacement and calculating the metrics for each resampled dataset. The 95\% confidence intervals (CIs) are then derived from the distribution of these bootstrapped estimates.

\subsection{Baselines}
We start with a very simple baseline of non-finetuned models--gemini-pro 1.0, GPT-4, and PaLM-2. We have summarized the results in Table \ref{tab:test_set_results}. We notice that all baseline models achieve a very high recall. Among the models that are not fine-tuned, the inference with GPT-4 results in best performance with a precision of 78.2\% (73.1, 83.2), f1-score of 0.845 (0.807, 0.881) and an accuracy of 94.0\% (91.2, 96.2).  The inference with the PaLM-2 base model achieves 59.8\% (54.7, 65.0) precision, 0.710 (0.667, 0.752) f1-score and 87.6\% (84.1, 90.7) accuracy. The inference with gemini-pro 1.0 results in a performance between PaLM-2 and GPT-4 with 65.7\% (60.6, 70.9) precision, 0.761 (0.719, 0.802)  f1-score and 90.0\% (86.8, 92.8) accuracy. In addition to these baselines, we also use baselines with naive combinations of gemini-pro 1.0 and PaLM-2 prediction (see Appendix \ref{naive_ensemble} for details.) 

\begin{table}[h]
\centering
\caption{Average Recall, Precision, F-1 Score, and Accuracy on the Test Set (Values reported as mean (95\% CI))}
\scalebox{.80}{
\begin{tabular}{llcccc} 
\toprule
\textbf{Model Type} & \textbf{Model/Training Data} & \textbf{Recall \%} & \textbf{Precision \%} & \textbf{F-1} & \textbf{Accuracy \%} \\
\midrule
\multirow{3}{*}{Non-Fine-Tuned} & PaLM-2 & 97.3 & 59.8 & 0.710 & 87.6 \\
 &  & (94.9, 99.3) & (54.7, 65.0) & (0.667, 0.752) & (84.1, 90.7) \\ 
\cmidrule(lr){2-6}
 & GPT-4 (gpt-4-0613) & 98.6 & \textbf{78.2} & \textbf{0.845} & \textbf{94.0} \\
 &  & (97.1, 99.8) & (73.1, 83.2) & (0.807, 0.881) & (91.2, 96.2) \\
\cmidrule(lr){2-6}
 & Gemini Pro 1.0 & \textbf{99.4} & 65.7 & 0.761 & 90.0 \\
 & & (98.4, 100.0) & (60.6, 70.9) & (0.719, 0.802) & (86.8, 92.8) \\
\cmidrule(lr){1-6}
\multirow{5}{*}{Fine-Tuned PaLM-2} & Human Labels & 98.9 & \textbf{87.3} & \textbf{0.908} & \textbf{96.0} \\
 & & (97.3, 100.0) & (82.9, 91.4) & (0.875, 0.938) & (93.5, 98.0) \\
\cmidrule(lr){2-6} 
 & All Gemini labels & \textbf{100} & 63.8 & 0.744 & 88.1 \\
 & & (100.0, 100.0) & (58.1, 69.2) & (0.698, 0.788) & (84.2, 91.5) \\
\cmidrule(lr){2-6}
 & Filtered Gemini labels & 97.7 & 85.9 & 0.894 & 95.7 \\
 & & (95.5, 99.5) & (81.2, 90.3) & (0.858, 0.927) & (93.1, 97.7) \\
\cmidrule(lr){2-6}
 & Human + Gemini labels & 98.9 & 86.4 & 0.903 & 95.8 \\
 & & (97.3, 100.0) & (81.6, 90.7) & (0.867, 0.934) & (93.2, 97.8) \\
\bottomrule
\end{tabular}
}
\label{tab:test_set_results}
\end{table}

\subsection{Fine-tuned Models}
We conduct four sets of fine-tuning experiments using PaLM-2 models. We first conduct fine-tuning on PaLM-2 model using the non-expert human annotation obtained as described in section \ref{selective}. This serves as a strong benchmark for our remaining fine-tuning experiments that use gemini-pro 1.0 based annotations for fine-tuning PaLM-2. The second and third fine-tuning experiments  differ in the sense that while one of the experiments use entirety of  gemini-pro's labels during fine-tuning, the other experiment is fine-tuned on only those gemini-pro labels where there is a consensus between gemini-pro's inference for JSON and text-based representation of a given table. Finally, we explore a hybrid approach to further refine the fine-tuning process. Instead of discarding the samples lacking consensus between JSON and text-based inferences, we replace the Gemini Pro labels for those samples with human annotations. This results in a fine-tuning dataset consisting of 2518 Gemini-annotated samples and 282 human-annotated ones.

As seen in table \ref{tab:test_set_results}, fine-tuning PaLM-2 with labels generated by Gemini-pro 1.0 leads to improvements over the base PaLM-2 model's performance. However, this improvement is nuanced. When the model is fine-tuned using the entirety of gemini-pro's labels, it improves over the baseline PaLM-2 model. However, the precision, f1-score, and accuracy compared to the standalone gemini-pro 1.0 model remains inferior. Optimal fine-tuning is achieved through only incorporating labels for which there is a consensus between gemini-pro's JSON and text-based inferences for a table. This fine-tuned variant (table \ref{tab:test_set_results} second last row), not only exceeds the performance of the base PaLM-2 and gemini-pro 1.0 models, but also narrows the gap to the precision and f1-score of PaLM-2 fine-tuned with human labels. Our hybrid labeling strategy (table \ref{tab:test_set_results} last row), incorporating human annotations for cases lacking consensus, further bridges the gap with the model fine-tuned on human annotations.

Finally, our fine-tuned models surpass the performance of naive ensemble approaches combining gemini-pro 1.0 and PaLM-2, as detailed in Appendix \ref{naive_ensemble}. Yet, it is worth noting that the naive ensembles still offer better results than either standalone base model. These findings underscore the value of naive ensembles for preliminary analysis and for scenarios where fine-tuning isn’t feasible, while also highlighting the potentially superior results attainable with fine-tuned models.

\begin{figure}[t]
  \centering
  \begin{minipage}{0.48\textwidth}
    \centering
    \includegraphics[width=\textwidth]{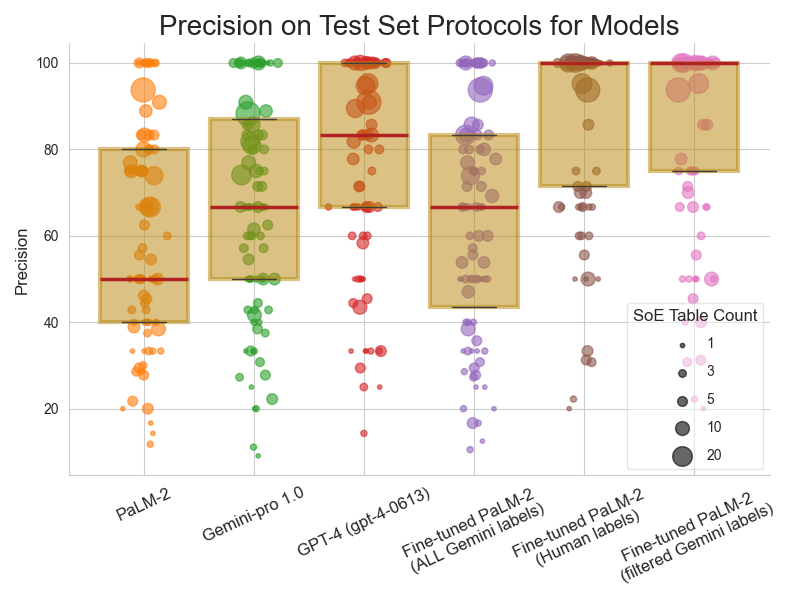} 
  \end{minipage}
  \hfill
  \begin{minipage}{0.48\textwidth}
    \centering
    \includegraphics[width=\textwidth]{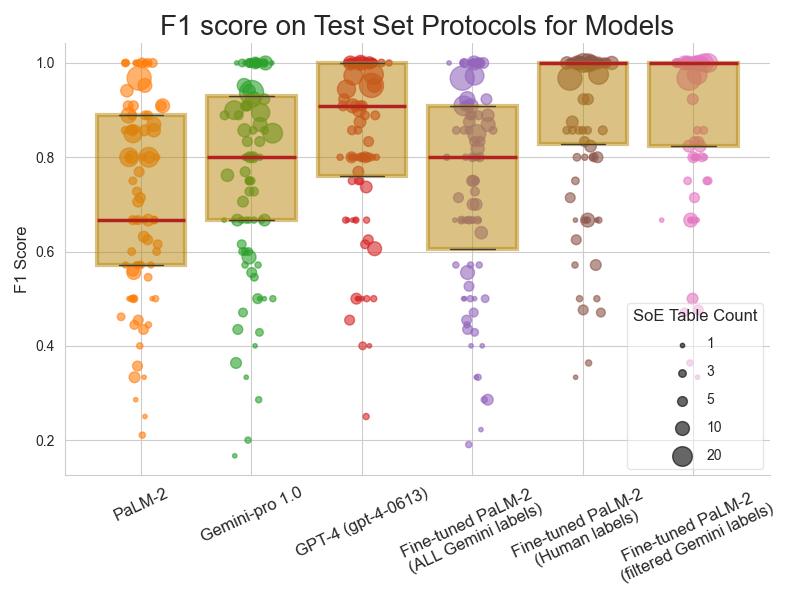}
  \end{minipage}
  \caption{\textbf{Precision and F1 Score across models for 91 protocols in test set} (hybrid approach not shown for brevity.) Bubble sizes represent the number of SoE tables within a protocol. PaLM-2 models fine-tuned with human labels and consensus-based gemini-pro 1.0 labels achieve a median precision of 100\% and F1 score of 1.}
  \label{fig:metrics}
\end{figure}

We plot the results for precision and f1-score in figure \ref{fig:metrics} corresponding to models in table \ref{tab:test_set_results} for all 91 test protocols. The bubble size in the scatter plots corresponding to the protocols are proportional to the number of SoE tables present in that protocol. We also overlay the boxplot on the scatter plot to show the the precision and f1-scores across individual protocols in the test set. We see that PaLM-2 models fine-tuned on human labels as well as consensus-based gemini-pro 1.0 labels achieve a median precision of 100\% and median f1-score of 1. This means that for at least 50\% of the protocols, the SoE table detection step in digitization workflow will be processed correctly without needing any further correction. 

\section{Discussion} 
Our study proposes a novel approach to fine-tuning Large Language Models for specialized domains where labels can be especially difficult to obtain. The approach of utilizing noisy labels from an LLM (gemini-pro) for fine-tuning another LLM (PaLM-2) demonstrates a scalable and cost-effective alternative to conventional expert annotation processes. Notably, the introduction of a label filtering mechanism, which selects labels for fine-tuning only when there is agreement between dual data representations (JSON and text) of a table, leads to substantial improvements in model performance. This method not only outperforms the base model and the label-generating LLM, it also approaches the performance level of models fine-tuned with human annotations on our table classification task, highlighting the potential of our fine-tuning strategy to effectively leverage auto-generated labels. The filtering mechanism also provides an easy way to identify potential samples that may require expert labeling. As demonstrated by our hybrid labeling approach, this can significantly reduce the annotation workload while potentially bringing fine-tuned models on par with models fine-tuned on human labels. This targeted approach to human annotation allows for a more efficient allocation of resources, maximizing the benefits of both automated and expert labeling. While we use JSON and text based consensus approach as a proxy for selecting high quality training labels for our specific table classification task, any multi-modal data representation can serve a similar purpose in filtering out potentially noisy labels.

Despite its strengths, our approach has several limitations. The task of SoE table detection is highly specialized and relies on data from our internal Clinical Trial Management System (CTMS) software, which may not fully capture the variety and complexity of clinical trial protocols encountered in broader applications. This specificity could limit the generalizability of our findings to other types of documents or on different table classification tasks. If LLMs perform poorly across the board on a specific task, automated generation of labels may not be feasible even with powerful models like gemini-pro. Moreover, while our study underscores the feasibility and effectiveness of using LLM-generated labels for fine-tuning, it lacks a direct comparison with a baseline model fine-tuned on expert annotations due to the high costs and resource requirements associated with obtaining such annotations. This comparison could have provided a clearer benchmark for evaluating the relative performance of our approach. Importantly, the reliance on auto-label generation and consensus-based fine-tuning may introduce or perpetuate biases inherent in the models used for label generation. Depending on specific context and fine-tuning task, these biases may manifest as demographic, entity, or domain-specific biases, affecting the accuracy and fairness of the fine-tuned model, particularly in sensitive domains like healthcare.
While our hybrid approach, incorporating human annotations for challenging or low-confidence cases, can help mitigate some of these biases, comprehensive fairness and bias evaluation specific to both the auto-labeling and fine-tuning steps are still essential for detecting and mitigating biases in auto-generated labels and fine-tuned models. This can pave the way for broader adoption of LLMs in data-rich, but expert-scarce domains like healthcare, where processes often rely heavily on manual workflows.

\section{Data \& Code Availability}
Due to the terms of our data sharing agreement, we are unable to provide access to the dataset. Additionally, the code used in this study is part of our proprietary software and cannot be shared. 

\section{Author Contribution}
BK and YJ conceived of utilizing LLM-based labels for fine-tuning. JA, EY, and BK conceived of the initial fine-tuning experiments using human labeled data. NL came up with the idea and initial prompt to use an LLM for table classification task. BK conducted all the experiments and wrote the draft of the paper. All authors reviewed the draft and edited the manuscript and take responsibility for all aspects of the work.

\section{Ethics Declaration}
\subsection{Competing interests}
All authors are employees and shareholders of Verily Life Sciences LLC.





\bibliography{references}

\newpage
\appendix
\section{Schedule of Event Tables}\label{soe_table}
In clinical trials, a schedule of events table is a table outlining the timeline and sequence of assessments, procedures, and data collection that will take place during the study. This table is an important part of the study protocol and provides a comprehensive overview of the study activities for both the researchers and participants. The schedule of events table typically includes the following information:
\bigskip
\begin{enumerate}
    \item \textbf{Study Visits:} 
This includes different study visits or assessment time-points, such as screening, baseline, treatment periods, follow-up visits, and the end of study. Typically the timing of each visit (e.g., day, week, month) are also specified.

    \item \textbf{Assessments and Procedures:}
The Schedule-of-events table also describes the various assessments, tests, and procedures that will be performed at each study visit. This may include informed consent, physical examinations, vital sign measurements, laboratory tests, imaging studies, patient-reported outcomes, and any other relevant data collection.

    \item \textbf{Data Collection:}
The table includes the data that will be collected at each study visit, such as adverse events, concomitant medications, and any other relevant information.
\end{enumerate}

We provide a sample SoE table (refer table \ref{tab:visit-schedule}) based on \href{https://www.nia.nih.gov/sites/default/files/2019-10/startup_protocol_template_09202019.docx}{NIH template} and two non-SoE tables (refer tables \ref{tab:followup-schedule} and \ref{tab:pk-ig-schedule}) below. The first table has clear screening, treatment and follow-up period. It specifies various visit with time information as well as window during which the visit can take place. The second table looks like an SoE table in terms of structure, but it doesn’t have clearly demarcated screening, treatment and follow-up. Further, it only specifies specific lab tests at the start of the diagnosis and completion of therapy and lacks treatment period information. Often this table would require protocol digitization specialists to look at additional context (like surrounding texts on the page) in the protocol to determine whether or not it is a SoE table. The last table specifies pharmacokinetic collections and is not a SoE table (see prompts in appendix \ref{prompts} which we wrote in consultation with the digitizers for SoE tables)

Note that we are unable to provide identical sample tables from our own dataset due to limitations on data sharing. The first example of the SoE table is taken from the NIH template for SoE tables and the last two tables are fictitious and are not from any actual clinical trial protocols.
\clearpage

\subsection{Example SoE and Non-SoE Tables}\label{example_soe}
\begin{table}[h]
\caption{Sample SoE Table}
\resizebox{\textwidth}{!}{
\begin{tabular}{l ccccccc}
\toprule
\multirow{2}{*}{Assessment} & \multicolumn{1}{c}{Screening:} & \multicolumn{5}{c}{Treatment Visits} & \multicolumn{1}{c}{Follow-up} \\ 
\cmidrule(lr){2-2} \cmidrule(lr){3-7} \cmidrule(l){8-8}
 & Visit (Day -14 to -1) & Baseline, & Visit 2 & Visit 3 & Visit 4 & Visit 5  & Final \\
 & & Enrollment, &  &  &  &  & Visit \\
 & & Visit 1 (Day 0) & (Day 7$\pm$2 Days) & (Day 14$\pm$2 Days) & (Day 21$\pm$2 Days) & (Day 28$\pm$2 Days) &  (Day 70$\pm$7 Days) \\
\midrule
Informed Consent Form & X &  &  &  &  &  &  \\
Demographics & X &  &  &  &  &  & X \\
DXA & X &  &  &  &  &  & X \\
Medical History & X &  &  &  &  &  &  \\
General Physical Examination & X & X & X & X & X &  & X \\
Current Medications & X & X &  &  &  &  &  \\
Blood Chemistries & X & X & X & X & X & X & X \\
Hematology & X & X & X & X & X & X & X \\
Urine Analysis & X & X & X & X & X & X &  \\
Vital Signs & X & X & X & X & X & X & X \\
Inclusion/Exclusion Criteria & X & X &  &  &  &  &  \\
Enrollment/Randomization &  & X &  &  &  &  & X \\
Treatment Administration Form &  & X & X & X & X & X & X \\
Concomitant Medications &  & X & X & X & X & X &  \\
Adverse Events &  & X & X & X & X & X & X \\
\bottomrule
\end{tabular}}
\label{tab:visit-schedule}
\end{table}
\begin{table}[htbp]
\caption{Non-SoE Table Example 1}
\resizebox{\textwidth}{!}{
\begin{tabular}{l cccccc}
\toprule
\multirow{2}{*}{Evaluation} &  & \multicolumn{4}{c}{Months Following the Completion of Therapy} &  \\
\cmidrule(lr){3-6} 
 & Diagnosis & 3 & 9 & 24 & 48 & \\
\midrule
Physical measurements & X &  & X & X &  X &  \\
IGF-1 & X &  & X & X &  &  \\
TSH & X & X & X & X &  &  \\
Morning Cortisol (7AM-9AM) & X &  &  & X &  &  \\
Systolic BP & X & X & X & X & X &  \\
Serum Sodium & X &  & X & X &  &  \\
HbA1c & X &  & X & X &  X &  \\
Serum Calcium & X &  &  & X &  &  \\
\bottomrule
\end{tabular}}
\label{tab:followup-schedule}
\end{table}

\begin{table}[htbp]
\caption{Non-SoE Table Example 2}
\resizebox{\textwidth}{!}{
\begin{tabular}{l c c c c c c}
\toprule
Assessment or Procedure & Dose & Day & Time & Time Window & Pharmacokinetics & Immunogenicity \\
\midrule
 & Dose 1 & Day 2 & Pre-dose &  & X & X \\
 & Dose 2 & Day 9 & Pre-dose & $\pm$2 hours & X &  \\
Cycle 1 & Dose 3 & Day 16 & Pre-dose &  &  & X \\
 & Dose 4 & Day 25 & Pre-dose & $\pm$8 hours & X &  \\
Cycle 2 & Dose 5 & Day 30 & Pre-dose &  &  &  \\
\midrule
Final Assessment &  & End of Tx &  &  &  &  \\
\midrule
Follow-up Review &  & Day 30 & Post 5 weeks &  & X & X \\
\bottomrule
\end{tabular}}
\label{tab:pk-ig-schedule}
\end{table}
\section{Prompts}\label{prompts}
\subsection{Prompt for JSON based inference}
We use the following prompt for inference with JSON representation of the table:
\begin{verbatim}
A schedule of events or activities (SoE or SoA) table in a clinical
trial protocol specifies a plan of care for participants. 
Here are some characteristics of SoE/SoA tables:
1. The header rows specify the name and timing of a series of visits
to the research site where the participants receive some assessments
or treatments.
2. The visits are usually arranged in three phases: screening visits,
treatment visits and follow-up visits. A typical SoE or SoA table
includes the visits of ALL the three phases or periods.
3. Body rows indicating the occurrence of an assessment or treatment
during specific visits, often denoted by symbols like 'X', '✓', or '•'. 
Some cells may have additional textual specifications. Key terms often
found in an SoE or SoA table include: "Informed Consent", 
"randomization", "treatment", "protocols", and "timing of visit".
If you find these keywords (especially "Informed Consent"), this 
indicates an SoE or SoA table.
Following tables are NOT SoE or SoA tables:
    1. An SoE or SoA table is NOT a table describing the timepoints
    when a specific assessment should be performed, such as a table
    specific to laboratory assessments, pharmacokinetic collections,
    or pharmacodynamic collections. These will often break down an
    assessment into hourly collections after an occurrence, like a
    pharmacokinetic collection that is performed many times on a 
    single day in relation to treatment administration (0h post-dose,
    2h post-dose, 6h post-dose, and so on). These are supplemental
    tables that greatly expand upon an abbreviated description in
    the SoE, but are NOT an SoE table.
    2. An SoE or SoA table is NOT a document history table listing
    all previous protocol versions that have been amended and a 
    summary of their changes.
    3. An SoE or SoA table is NOT an objectives table, describing 
    the research and statistical goals of the research study 
    (also endpoints, outcomes, etc.)
    4. An SoE or SoA table is NOT a table describing adequate organ
    function or laboratory values
    5. An SoE or SoA table is NOT a table describing dose 
    modifications and toxicity in regards to the research treatment
Given the input as a table in the JSON format, return YES if it is 
an SoE or SoA table or return NO if it is not. Do not output anything
else.
Input table in JSON format:
{table}
Your answer (One of YES or NO):
\end{verbatim}
\subsection{Prompt for text based inference}
The text based model inference for table classification is done with the following prompt:
\begin{verbatim}
A schedule of events or activities (SoE or SoA) table in a clinical trial
protocol specifies a plan of care for participants. Here are some 
characteristics of SoE/SoA tables:
1. The header rows specify the name and timing of a series of visits
to the research site where the participants receive some assessments
or treatments.
2. The visits are usually arranged in three phases: screening visits,
treatment visits and follow-up visits. A typical SoE or SoA table 
includes the visits of ALL the three phases or periods.
3. Body rows indicating the occurrence of an assessment or treatment
during specific visits, often denoted by symbols like 'X', '✓', or '•'.
Some cells may have additional textual specifications.
Key terms often found in an SoE or SoA table include: "Informed Consent", 
"randomization", "treatment", "protocols", and "timing of visit". 
If you find these keywords (especially "Informed Consent"), 
this indicates an SoE or SoA table.
Following tables are NOT SoE or SoA tables:
    1. An SoE or SoA table is NOT a table describing the timepoints
    when a specific assessment should be performed, such as a table 
    specific to laboratory assessments, pharmacokinetic collections, 
    or pharmacodynamic collections. These will often break down an 
    assessment into hourly collections after an occurrence, like a 
    pharmacokinetic collection that is performed many times on a 
    single day in relation to treatment administration (0h post-dose, 
    2h post-dose, 6h post-dose, and so on). These are supplemental 
    tables that greatly expand upon an abbreviated description in the SoE, 
    but are NOT an SoE table.
    2. An SoE or SoA table is NOT a document history table listing all 
    previous protocol versions that have been amended and a summary of
    their changes.
    3. An SoE or SoA table is NOT an objectives table, describing the 
    research and statistical goals of the research study (also endpoints,
    outcomes, etc.)
    4. An SoE or SoA table is NOT a table describing adequate organ 
    function or laboratory values
    5. An SoE or SoA table is NOT a table describing dose modifications
    and toxicity in regards to the research treatment
One way to identify an SoE or SoA table is to look at the text outside
the table. Specifically, look for terms  like Schedule of Events, 
Schedule of Assessment, Schedule of Activities, Study Calender,
Study Parameters, Study Schedule and related terms.
If you see any of these terms or related terms in the text data you can
conclude that it indicates an SoE or SoA table.
If you don't see any of these terms in the text data you should look 
at the whole text data to determine if it is an SoE or SoA table.
Your goal is to determine if the provide text data is from an SoE or 
SoA table or not. The text data includes all the text before, inside 
and after the table.
Return YES if it is an SoE or SoA table or return NO if it is not. 
Do not output anything else.
Text Data (including before, inside and after the table):
{text}
Your answer (YES or NO):
\end{verbatim}
\clearpage
\section{Naive Combination of Gemini-pro and PaLM-2 models}\label{naive_ensemble}

To broaden our baseline comparisons, we experimented with a naive ensemble approach by combining the outputs of the gemini-pro 1.0 and PaLM-2 models. This exploratory analysis aimed to assess whether a naive combination of model inferences could leverage the strengths of both individual models to improve the detection of Schedule-of-Event (SoE) tables.

Our ensemble strategy entailed aggregating predictions from both models, each producing two sets of inferences for the tables in clinical trial protocols based on JSON and text representations. We established varying thresholds—from a minimum of one to a maximum of four affirmative (``YES") inferences—to determine when a table should be classified as a SoE. The performance metrics of the naive ensemble models, detailed in Table \ref{tab:model-performance}, indicate that the ensemble outperforms the individual models when a threshold of at least two affirmative inferences is applied. This specific threshold represents a balance, capturing the consensus across the models while mitigating the impact of any one model's false positives or negatives.  Nonetheless, the performance of naive ensemble approaches remained inferior to the fine-tuned models (PaLM-2 fine-tuned with human labels or gemini annotated and consensus-filtered labels) at all thresholds underscoring the value of fine-tuning over simple ensemble methods in this context. The results of our naive ensemble models show that while aggregation techniques can yield benefits, they are outperformed by a more sophisticated method of fine-tuning models with carefully curated labels.

\begin{table}[h]
\caption{Performance of models when using various thresholds for classifying as SoE Tables}
\centering
\begin{tabular}{l|c|c|c|c}
\toprule
\textbf{Model} & \textbf{Recall} & \textbf{Precision} & \textbf{F-1 Score} & \textbf{Accuracy} \\
\midrule
PaLM-2-Gemini Naive & 100\% & 51.5\% & 0.65 & 83.6\% \\
Ensemble-1 &  &  &  &  \\
(SoE if at least one inference is SoE) & & & & \\
\midrule
PaLM-2-Gemini Naive & 99.5\% & 75.6\% & 0.83 & 92.6\% \\
Ensemble-2 &  &  &  &  \\
(SoE if $>= 2$ inferences are SoE) & & & & \\
\midrule
PaLM-2-Gemini Naive & 94.9\% & 83.6\% & 0.86 & 94.9\% \\
Ensemble-3 &  &  &  &  \\
(SoE if $>= 3$ inferences are SoE) & & & & \\
\midrule
PaLM-2-Gemini Naive & 87.2\% & 86.2\% & 0.85 & 95.1\% \\
Ensemble-4 &  &  &  &  \\
(SoE if all inferences are SoE) & & & & \\
\bottomrule
\end{tabular}
\label{tab:model-performance}
\end{table}
\clearpage
\section{Additional Recall and Precision Metrics}\label{additional_metrics}
\begin{table}[ht]
\caption{Additional Recall and Precision Metrics}
\centering
\begin{tabular}{l|c|c|c|c}
\toprule
Model & \begin{tabular}[c]{@{}c@{}}\% of protocol \\ with $>$60\% \\ precision\end{tabular} & \begin{tabular}[c]{@{}c@{}}\% of protocol \\ with $>$80\% \\ precision\end{tabular} & \begin{tabular}[c]{@{}c@{}}\% of protocol \\ with 100\% \\ precision\end{tabular} & \begin{tabular}[c]{@{}c@{}}\% of protocol \\ with 100\% \\ recall\end{tabular} \\
\midrule
PaLM-2 & 44.0 & 22.0 & 14.3 & 93.4 \\
GPT-4 (gpt-4-0613) & 75.8 & 56.0 & 42.9 & 95.6 \\ 
Gemini Pro 1.0 & 56.0 & 33.0 & 22.0 & 98.9 \\
\midrule
Fine-tuned PaLM & \multirow{2}{*}{\textbf{85.7}} & \multirow{2}{*}{\textbf{71.4}} & \multirow{2}{*}{\textbf{68.1}} & \multirow{2}{*}{97.8} \\
(Using Human Labels) & & & & \\ 
\midrule
Fine-tuned PaLM-2 & \multirow{2}{*}{53.8} & \multirow{2}{*}{28.6} & \multirow{2}{*}{19.8} & \multirow{2}{*}{\textbf{100.0}} \\
(using ALL Gemini & & & & \\
Labels) & & & & \\
\midrule
Fine-tuned PaLM-2 & \multirow{2}{*}{82.4} & \multirow{2}{*}{69.2} & \multirow{2}{*}{64.8} & \multirow{2}{*}{95.0} \\
(Using Filtered Gemini & & & & \\
Labels) & & & & \\
\bottomrule
\end{tabular}
\label{tab:additional-metrics}
\end{table}






\end{document}